# Image-Grounded Conversations: Multimodal Context for Natural Question and Response Generation

**Nasrin Mostafazadeh**[1*], **Chris Brockett**[2], **Bill Dolan**[2], **Michel Galley**[2], **Jianfeng Gao**[2],
**Georgios P. Spithourakis**[3*], **Lucy Vanderwende**[2]

1 University of Rochester,    2 Microsoft,

3 University College London

nasrinm@cs.rochester.edu,   chrisbkt@microsoft.com

## Abstract

The popularity of image sharing on social media and the engagement it creates between users reflects the important role that visual context plays in everyday conversations. We present a novel task, Image-Grounded Conversations (IGC), in which natural-sounding conversations are generated about a shared image. To benchmark progress, we introduce a new multiple-reference dataset of crowd-sourced, event-centric conversations on images. IGC falls on the continuum between chit-chat and goal-directed conversation models, where visual grounding constrains the topic of conversation to event-driven utterances. Experiments with models trained on social media data show that the combination of visual and textual context enhances the quality of generated conversational turns. In human evaluation, the gap between human performance and that of both neural and retrieval architectures suggests that multi-modal IGC presents an interesting challenge for dialogue research.

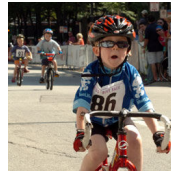

**User1:** My son is ahead and surprised!
**User2:** Did he end up winning the race?
**User1:** Yes he won, he can't believe it!

Figure 1: A naturally-occurring Image-Grounded Conversation.

## 1 Introduction

Bringing together vision & language in one intelligent conversational system has been one of the longest running goals in AI (Winograd, 1972). Advances in image captioning (Chen et al., 2015; Fang et al., 2014; Donahue et al., 2014; Chen et al., 2015) have enabled much interdisciplinary research in vision and language, from video transcription (Rohrbach et al., 2012; Venugopalan et al., 2015), to answering questions about images (Antol et al., 2015; Malinowski and Fritz, 2014), to storytelling around series of photographs (Huang et al., 2016).

A majority of the recent work on vision & language focuses on either describing (captioning) the image or answering questions about their visible content. Observing how people naturally engage with one another around images in social media, it is evident that it is often in the form of conversational threads. On Twitter, for example, uploading a photo with an accompanying tweet has become increasingly popular: as of June 2015, 28% of tweets contain an image (Morris et al., 2016). Across social media, the conversations around shared images are beyond what is explicitly visible in the image. Figure 1 illustrates such a conversation. As this example shows, the conversation is grounded not only in the visible objects (e.g., the boys, the bikes) but more importantly, in the events and actions (e.g., the race, winning) implicit in the image which is accompanied by the textual utterance. To humans, it is these latter aspects that are likely to be the most interesting and meaningful components of a natural conversation, and to the systems, inferring such implicit aspects can be the most challenging.

In this paper we shift the focus from image as an artifact (as is in the existing vision & language work, to be described in Section 2), to image as the context for interaction: we introduce the task of Image-Grounded Conversation (IGC) in which a system must generate conversational turns to proactively drive it forward. IGC thus falls on a continuum between chit-chat (open-ended) and goal-oriented task-completion dialog systems, where the visual context in IGC naturally serves as a detailed topic for a conversation. As conversational agents gain increasing ground in commercial settings (such as Siri, Alexa, etc.), these agents will increasingly need to engage humans in ways that seem intelligent and anticipatory of future needs. For example, a conversational agent might engage in a conversation with a user about a camera-roll image to seek to elicit background information from the user (e.g., special celebrations, favorite food, the name of the friends and family, etc.).

This paper draws together two threads of investigation that have hitherto remained largely unrelated: vision & language and data-driven conversation modeling. Its contributions are threefold: (1) we introduce multimodal conversational context for formulating questions and responses around images. We support benchmarking on the task with a high-quality, crowd-sourced dataset of 4,222 multi-turn multi-reference conversations grounded on event-centric images (that will be publicly released). We analyze various characteristics of this IGC dataset in Section 3.1. (2) We investigate the application of deep neural generation and retrieval approaches for question and response generation tasks (Section 5), trained on 250K 3-turn naturally-occurring image-grounded conversations found on Twitter. (3) Our experiments suggest that the combination of visual and textual context improves the quality of generated conversational turns (Section 6-7). It is our hope that the introduction of this task will spark a new interest in multimodal conversation modeling.

## 2 Related Work

### 2.1 Vision and Language

Visual features combined with language modeling have shown good performance both in image captioning (Devlin et al., 2015; Xu et al., 2015; Fang et al., 2014; Donahue et al., 2014) and in question answering on images (Antol et al., 2015; Ray et al., 2016; Malinowski and Fritz, 2014), when trained on large datasets, such as the COCO dataset (Lin et al., 2014). In Visual Question Answering (VQA) (Antol et al., 2015), a system is tasked with answering a question about a given image, where the questions are constrained to be answerable directly from the image. In other words, the VQA task primarily serves to evaluate the extent to which the system has recognized the explicit content of the image.

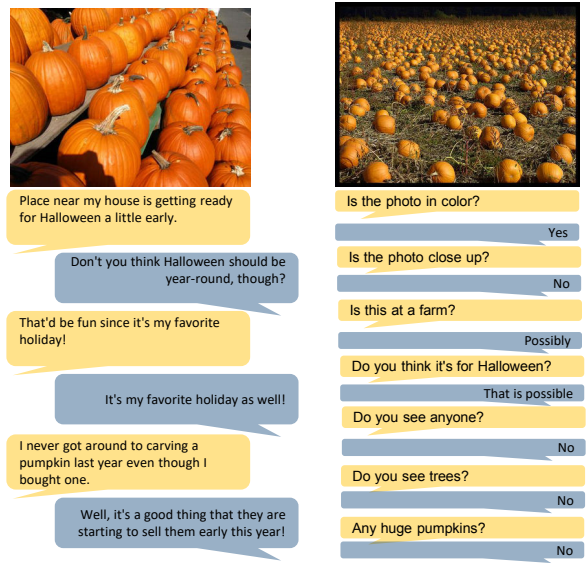

Figure 2: An example crowdsourced conversation for the task of IGC (left) and VisDial (right).

Das et al. (2017) extend the VQA scenario by collecting sequential questions from people who are shown only an automatically generated caption, not the image itself. The utterances in this dataset, called 'Visual Dialog' (VisDial), are best viewed as simple one-sided QA exchanges in which humans ask questions and the system provide answers. Figure 2 contrasts an example ICG conversation with the VisDial dataset. As this example shows, IGC involves natural conversations with the image as the grounding, where the literal objects (e.g., the pumpkins) may not even be mentioned in the conversation at all, whereas VisDial targets explicit image understanding.

Mostafazadeh et al. (2016b) introduce the task of visual question generation (VQG), in which the system itself outputs questions about a given image. Questions are required to be 'natural and engaging', i.e. a person would find them interesting to answer, but need not be answerable from the image alone. In this work, we introduce multimodal context, recognizing that images commonly come associated with a verbal commentary that can affect the interpretation. In addition, we include conversational response generation within the purview of the task.

### 2.2 Data-Driven Conversational Modeling

This work is also closely linked to research on data-driven conversation modeling. Ritter et al. (2011) posed response generation as a machine translation task, learning conversations from parallel message-response pairs found on social media. Their work has been successfully extended with the use of deep neural models (Sordoni et al., 2015; Shang et al., 2015; Serban et al., 2015a; Vinyals and Le, 2015; Li et al., 2016a,b). Sordoni et al. (2015) introduce a context-sensitive neural language model that selects the most probable response conditioned on the conversation history (i.e., a text-only context). In this paper, we extend the contextual approach with the addition of multimodal features to build models that are capable of not only responding but also asking questions on topics of interests to a human, which might allow a conversational agent to proactively drive the conversation forward.

## 3 Image-Grounded Conversations

### 3.1 Task Definition

We define the current scope of IGC as the following two consecutive conversational steps:

- **Question Generation:** Given a visual context $I$ and a textual context $T$ (e.g., the first statement in Figure 1), generate a coherent, natural question $Q$ about the image as the second utterance in the conversation. As seen in Figure 1, the question is not directly answerable from the image. Here we emphasize on questions as a way of potentially engaging a human in continuing the conversation.
- **Response Generation:** Given a visual context $I$, a textual context $T$, and a question $Q$, generate a coherent, natural, response $R$ to the question as the third utterance in the conversation. In the interest of feasible multi-reference evaluation, we pose question and response generation as two separate tasks. However, all the models presented in this paper can be fed with their own generated question to generate a response.

### 3.2 The ICG Dataset

The majority of the available corpora for developing data-driven dialogue systems contain task-oriented and goal-driven conversational data (Serban et al., 2015b). For instance, the Ubuntu dialogue corpus (Lowe et al., 2015) is the largest corpus of dialogues (almost 1 million mainly 3-turn dialogues) for the specific topic of troubleshooting Ubuntu problems. On the other hand, for chit-chat (open-ended) conversation modeling, which has become a high demand application in AI, there is no shared dataset in the community for meaningful tracking of the progress. Our IGC task nicely lies in between these two, where the visual grounding of event-centric images constrains the topic of conversation to event-rich and contentful utterances.

To enable benchmarking of progress in the IGC task, we constructed the IGC$_{Crowd}$ dataset for validation and testing purposes. We first sampled eventful images from the VQG dataset (Mostafazadeh et al., 2016b) which has been extracted by querying a search engine using event-centric query terms. These were then served in a photo gallery of a crowdsourcing platform we developed using the Turkserver toolkit (Mao et al., 2012), which enables synchronous and real-time interactions between crowd workers on Amazon Mechanical Turk (Mturk). Multiple workers wait in a virtual lobby to be paired with another worker who will be their conversation partner. After being paired, one of the users selects an image from the large photo gallery, after which the two users enter a chat window in which they have a short conversation about the selected image. We prompted the workers to naturally drive the conversation forward without using informal/IM language. To enable multi-reference evaluation (Section 6), we crowd-sourced five additional questions and responses for the IGC$_{Crowd}$ contexts and initial questions.

Table 2 summarizes basic dataset statistics. Table 1 shows three full conversations found in the IGC$_{Crowd}$ dataset. As the examples show, eventful images lead to conversations which are semantically rich and would seem to require commonsense reasoning. The IGC$_{Crowd}$ dataset will be publicly released to the research community.

In order to create the training data, we sampled 250K quadruples of {visual context, textual context, question, response} tweet threads from a larger dataset of 1.4 million threads, extracted from the Twitter Firehose. Twitter data is notoriously noisy. The details on cleaning the IGC$_{Twitter}$ dataset along with example conversations can be found in the supplementary material.

## 4 Task Characteristics

In this Section, we analyze the IGC dataset to highlight a range of phenomena specific to this task. For further statistical analysis of this dataset, please see the supplementary material.

| | | | |
|---|---|---|---|
| **Visual Context** | 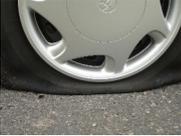 | 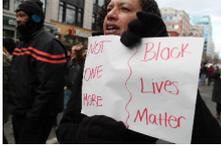 | 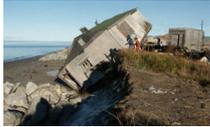 |
| **Textual Context** | This wasn't the way I imagined my day starting. | I checked out the protest yesterday. | A terrible storm destroyed my house! |
| **Question** | do you think this happened on the highway? | Do you think America can ever overcome its racial divide? | OH NO, what are you going to do? |
| **Response** | Probably not, because I haven't driven anywhere except around town recently. | I can only hope so. | I will go live with my Dad until the insurance company sorts it out. |
| **VQG Question** | What caused that tire to go flat? | Where was the protest? | What caused the building to fall over? |

Table 1: Example full conversations in our IGC$_{\text{Crowd}}$ dataset. For comparison, we also include VQG questions in which the image is the only context.

| IGC$_{\text{Crowd}}$ (val and test sets, split: 40% and 60%) | |
|---|---|
| # conversations = # images | 4,222 |
| total # utterances | 25,332 |
| # all workers participated | 308 |
| Max # conversations by one worker | 20 |
| Average payment per worker (min) | 1.8 dollars |
| Median work time per worker (min) | 10.0 |
| IGC$_{\text{Crowd}-multiref}$ (val and test sets, split: 40% and 60%) | |
| # additional references per question/response | 5 |
| total # multi-reference utterances | 42,220 |

Table 2: Basic Dataset Statistics.

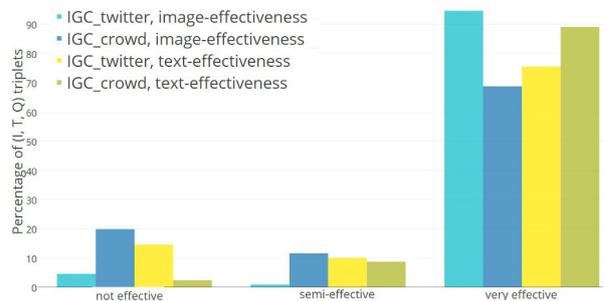

Figure 3: The effectiveness of textual and visual context for asking questions.

### 4.1 The Effectiveness of Multimodal Context

The task of IGC emphasizes modeling of not only visual but also textual context. We presented human judges with a random sample of 600 triplets of image, textual context, and question $(I, T, Q)$ from each IGC$_{\text{Twitter}}$ and IGC$_{\text{Crowd}}$ datasets and asked them to rate the effectiveness of the visual and the textual context. We define 'effectiveness' to be "the degree to which the image or text is required in order for the given question to sound natural". The workers were prompted to make this judgment based on whether or not the question already makes sense without either the image or the text. As Figure 3 shows, overall, both visual and textual contexts are indeed highly effective, and understanding both would be required for the question that was asked. We note that the crowd dataset more often requires understanding of the textual context than the Twitter set does, which reflects on its language-rich content.

### 4.2 Frame Semantic Analysis of Questions

The grounded conversations starting with questions in our datasets are full of stereotypical commonsense knowledge. To get a better sense of the richness of our IGC$_{\text{Crowd}}$ dataset, we manually annotated a random sample of 330 $(I, T, Q)$ triplets in terms of Minsky's Frames: Minsky defines 'frame' as follows: "*When one encounters a new situation, one selects from memory a structure called a Frame*" (Minsky, 1974). According to Minsky, a frame is a commonsense knowledge representation data-structure for representing stereotypical situations, such as a wedding ceremony. Minsky further connects frames to the nature of questions: "*[A Frame] is a collection of questions to be asked about a situation*". These questions can ask about the cause, intention, or side-effects of a presented situation.

We annotated[1] the FrameNet (Baker et al., 1998) frame evoked by the image $I$, to be called $(I_{FN})$, and the textual context $T$, $(T_{FN})$. Then, for the asked question, we annotated the frame slot $(Q_{FN-slot})$ associated with a context frame $(Q_{FN})$. For 17% of cases we could not find a corresponding $Q_{FN-slot}$ in FrameNet. As the exam-ple in Table 3 shows, the image in isolation often does not evoke any uniquely contentful frame, whereas the textual context frequently does. In only 14% of cases does $I_{FN}=T_{FN}$, which further supports the complementary effect of our multimodal contexts. Moreover, $Q_{FN}=I_{FN}$ for 32% our annotations, whereas $Q_{FN}=T_{FN}$ for 47% of the triplets, again, showing the effectiveness of rich textual context in determining the question to be asked.

---
[1]These annotations can be accessed through https://goo.gl/MVyGzP

| Visual Context | Textual Context | Question |
|---|---|---|
| 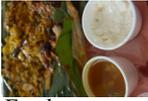 | Look at all this food I ordered! | Where is that from? |
| **FN** *Food* | *Request-Entity* | *Supplier* |

Table 3: FrameNet (FN) annotation of an example.

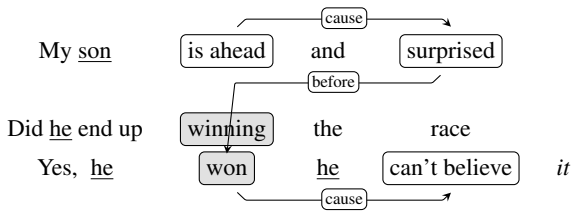

Figure 4: An example causal and temporal (CaTeRS) annotation on the conversation presented in Figure 1. The rectangular nodes show the event entities and the edges are the semantic links. For simplicity, we show the 'identity' relation between events using gray nodes. The coreference chain is depicted by the underlined words.

### 4.3 Event Analysis of Conversations

To further investigate the representation of events and any stereotypical causal and temporal relations between them in the IGC$_{Crowd}$ dataset, we manually annotated a sample of 20 conversations with their causal and temporal event structures. We followed the Causal and Temporal Relation Scheme (CaTeRS) (Mostafazadeh et al., 2016a) for event entity and event-event semantic relation annotations. Our analysis shows that the IGC utterances are indeed rich in events. On average, each utterance in IGC has 0.71 event entity mentions, such as 'win' or 'remodel'. The semantic link annotation reflects commonsense relation between event mentions in the context of the ongoing conversation. Figure 4 shows an example CaTeRS annotation. The distribution of semantic links in the annotated sample can be found in Figure 5. These statistics further suggest that in addition to jointly understanding the visual and textual context (including multimodal anaphora resolution, among other challenges), capturing causal and temporal relations between events is necessary for a system to successfully perform IGC task.

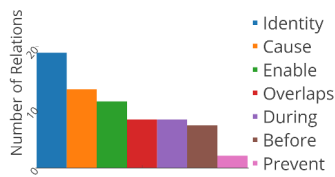

Figure 5: The frequency of event-event semantic links in a random sample of 20 IGC conversations.

## 5 Models

### 5.1 Generation Models

Figure 6 overviews our three generation models. Across all the models, we use the VGGNet architecture (Simonyan and Zisserman, 2014) for computing deep convolutional image features. We use the 4096-dimensional output of the last fully connected layer ($fc7$) as the input to all the models sensitive to visual context.

**Visual Context Sensitive Model** *(V-Gen)*. Similar to Recurrent Neural Network (RNN) models for image captioning (Devlin et al., 2015; Vinyals et al., 2015), *(V-Gen)* transforms the image feature vector to a 500-dimensional vector that serves as the initial recurrent state to a 500-dimensional one-layer Gated Recurrent Unit (GRU) which is the decoder module. The output sentence is generated one word at a time until the <EOS> (end-of-sentence) token is generated. We set the vocabulary size to 6000 which yielded the best results on the validation set. For this model, we got better results by greedy decoding. Unknown words are mapped to an <UNK> token during training, which is not allowed to be generated at decoding time.

**Textual Context Sensitive Model** *(T-Gen)*. This is a neural Machine Translation-like model that maps an input sequence to an output sequence (Seq2Seq model (Cho et al., 2014; Sutskever et al., 2014)) using an encoder and a decoder RNN. The decoder module is like the model described above, in this case the initial recurrent state being the 500-dimensional encoding of the textual context. For consistency, we use the same vocab size and number of layers as in the *(V-Gen)* model.

**Visual & Textual Context Sensitive Model** *(V&T-Gen)*. This model fully leverages both textual and visual contexts. The vision fea-ture is transformed to a 500-dimensional vector, and the textual context is likewise encoded into a 500-dimensional vector. The textual feature vector can be obtained using either a bag-of-words *(V&T.BOW-Gen)* representation, or an RNN *(V&T.RNN-Gen)*, as depicted in Figure 7. The textual feature vector is then concatenated to the vision vector and fed into a fully connected (FC) feed forward neural network. As a result, we obtain a single 500-dimensional vector encoding both visual and textual context, which then serves as the initial recurrent state of the decoder RNN.

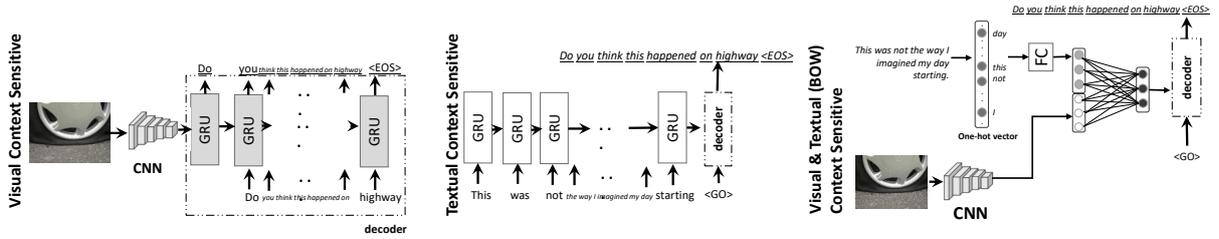

Figure 6: Question generation using the Visual Context Sensitive Model *(V-Gen)*, Textual Context Sensitive Model *(T-Gen)*, and the Visual & Textual Context Sensitive Model *(V&T.BOW-Gen)*, respectively.

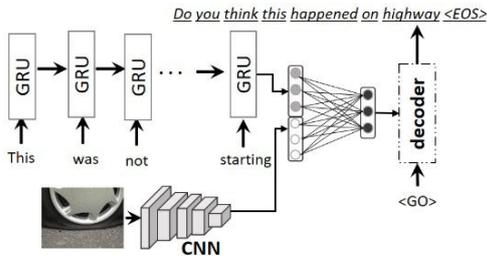

Figure 7: The visual & textual context sensitive model with RNN encoding *(V&T.RNN-Gen)*.

In order to generate the response (the third utterance in the conversation), we need to represent the conversational turns in the textual context input. There are various ways to represent conversational history, including a bag of words model, or a concatenation of all textual utterances into one sentence (Sordoni et al., 2015). For response generation, we implement a more complex treatment in which utterances are fed into an RNN one word at a time (Figure 7) following their temporal order in the conversation. An <UTT> marker designates the boundary between successive utterances.

**Decoding and Reranking.** For all generation models, at decoding time we generate the N-best lists using left-to-right beam search with *beam-size* 25. We set the maximum number of tokens to 13 for the generated partial hypotheses. Any partial hypothesis that reaches <EOS> token becomes a viable full hypothesis for reranking. The first few hypotheses on top of the N-best lists generated by Seq2Seq models tend to be very generic,[2] disregarding the input context. In order to address this issue we rerank the N-best list using the following score function:

$$log\ p(h|C) + \lambda\ idf(h,D) + \mu|h| + \kappa\ V(h) \quad (1)$$

where $p(h|C)$ is the probability of the generated hypothesis $h$ given the context $C$. The function $V$ counts the number of verb POS in the hypothesis and $|h|$ denotes the number of tokens in the hypothesis. The function *idf* is the inverse document frequency, simply computing how common a hypothesis is across all the generated N-best lists. Here $D$ is the set of all N-best lists and $d$ is a specific N-best list. We define $idf(h, D) = log \frac{|D|}{|\{d \in D: h \in d\}|}$, where we set $N$=10 to cut short each N-best list. These parameters were proven the most useful on reranking on the validation set. We optimize all the parameters of the scoring function towards maximizing the smoothed-BLEU score (Lin and Och, 2004) using the Pairwise Ranking Optimization algorithm (Hopkins and May, 2011).

### 5.2 Retrieval Models

In addition to generation, we implemented two retrieval models customized for the tasks of question and response generation. Work in vision and language has demonstrated the effectiveness of retrieval models, where one uses the annotation (e.g., caption) of a nearest neighbor in the training image set to annotate a given test image (Mostafazadeh et al., 2016b; Devlin et al., 2015; Hodosh et al., 2013; Ordonez et al., 2011; Farhadi et al., 2010).

**Visual Context Sensitive Model *(V-Ret)*.** This model only uses the given image for retrieval. First, we find a set of $K$ nearest training images for the given test image based on cosine similarity of the $fc7$ vision feature vectors. Then we retrieve those $K$ annotations as our pool of $K$ candidates. Finally, we compute the textual similarity among the questions in the pool according to a Smoothed-BLEU (Lin and Och, 2004) similarity score, then emit the sentence with the highest similarity to the rest of the pool.

---
[2]An example generic question is *where is this?* and a generic response is *I don't know*.

| | | Visual Context | 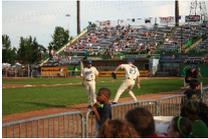 | 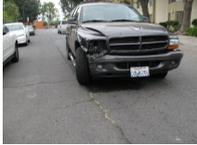 | 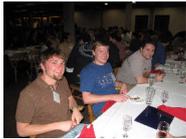 |
|---|---|---|---|---|---|
| Question Generation | | **Textual Context** | The weather was amazing at this baseball game. | I got in a car wreck today! | My cousins at the family reunion. |
| | | **Gold Question** | Nice, which team won? | Did you get hurt? | What is the name of your cousin in the blue shirt? |
| | | **V&T-Ret** | U at the game? or did someone take that pic for you? | **You driving that today?** | **U had fun?** |
| | | **V-Gen** | Where are you? | Who's is that? | Who's that guy? |
| | | **V&T-Gen** | **Who's winning?** | **What happened?** | **Where's my invite?** |
| Response Generation | | **Textual Context** | The weather was amazing at this baseball game. `<UTT>` Nice, which team won? | I got in a car wreck today! `<UTT>` Did you get hurt? | My cousins at the family reunion. `<UTT>` What is the name of your cousin in the blue shirt? |
| | | **Gold Response** | My team won this game. | No it wasn't too bad of a bang up. | His name is Eric. |
| | | **V&T-Ret** | 10 for me and 28 for my dad. | **Yes.** | lords cricket ground . beautiful. |
| | | **V&T-Gen** | ding ding ding! | **Nah, I'm at home now.** | **He's not mine!** |

Table 4: Example question and response generations on IGC$_{\text{Crowd}}$ test set. All the generation models use beam search with reranking. In the textual context, `<UTT>` separates different utterances. The generations in bold are acceptable utterances given the underlying context.

**Visual & Textual Context Sensitive Model (V&T-Ret).** This model uses a linear combination of $fc7$ and word2vec feature vectors for retrieving similar training instances.

## 6 Evaluation Setup

We provide both human (Table 5) and automatic (Table 6) evaluations for our question and response generation tasks on the IGC$_{\text{Crowd}}$ test set. We crowdsource our human evaluation on an AMT-like crowdsourcing system, asking seven crowd workers to each rate the quality of candidate questions or responses on a three-point Likert-like scale, ranging from 1 to 3 (the highest). To ensure a calibrated rating, we show the human judges all system hypotheses for a particular test case at the same time. System outputs were randomly ordered to prevent judges from guessing which systems were which on the basis of position. After collecting judgments, we averaged the scores throughout the test set for each model. We discarded any annotators whose ratings varied from the mean by more than 2 standard deviations.

Although human evaluation is to be preferred, and currently essential in open-domain generation tasks involving intrinsically diverse outputs, it is useful to have an automatic metric for day-to-day evaluation. For ease of replicability, we use the standard Machine Translation metric, BLEU (Papineni et al., 2002), which captures n-gram overlap between hypotheses and multiple references. Results reported in Table 6 employ BLEU with equal weights up to 4-grams at corpus-level on the multi-reference IGC$_{\text{Crowd}}$ test set. Although Liu et al. (2016) shows that BLEU fails to correlate with human judgment at the sentence level, correlation increases when BLEU is applied at the document or corpus level (Galley et al., 2015; Przybocki et al., 2008).

## 7 Experimental Results

We experiment with all the models presented in Section 5. For question generation, we use a visual & textual sensitive model that uses bag-of-words (V&T.BOW-Gen) to represent the textual context, which achieved better results. Earlier vision & language work such as VQA (Antol et al., 2015) has shown that a bag-of-words baseline outperforms LSTM-based models for representing textual input when visual features are available (Zhou et al., 2015). In response generation, which needs to account for textual input consisting of two turns, we use the *V&T.RNN-Gen* model as the visual & textual-sensitive model for the response rows of tables 5 and 6. Since generating a response solely from the visual context is unlikely to be successful, we do not use the *V-Gen* model in response generation. All the models are trained on IGC$_{\text{Twitter}}$ dataset, except for the model labeled VQG, which shares the same architecture as with the *(V-Gen)* model, but is trained on 7,500 questions from the VQG dataset (Mostafazadeh et al., 2016b) as a point of reference. We also include the gold human references from the IGC$_{\text{Crowd}}$ dataset in the human evaluation to achieve a bound on human performance.

|  | **Human** | **Generation** (Greedy) | | | **Generation** (Beam, best) | | | | **Generation** (Reranked, best) | | | **Retrieval** | |
|  | Gold | Textual | Visual | V & T | Textual | Visual | V & T | VQG | Textual | Visual | V & T | Visual | V & T |
| --- | --- | --- | --- | --- | --- | --- | --- | --- | --- | --- | --- | --- | --- |
| Question | <u>2.68</u> | 1.46 | 1.58 | 1.86 | 1.07 | 1.86 | **2.28** | 2.24 | 1.03 | 2.06 | 2.13 | 1.59 | 1.54 |
| Response | <u>2.75</u> | 1.24 | – | 1.40 | 1.12 | – | **1.49** | – | 1.04 | – | 1.44 | – | 1.48 |

Table 5: Human judgment results on the IGC$_{\text{Crowd}}$ test set. The maximum score is 3. Per model, the human score is computed by averaging across multiple images. The boldfaced numbers show the highest score among the systems. The overall highest scores (underlined) are the human gold standards.

|  | | **Generation** | | | **Retrieval** | |
|  | Textual | Visual | V & T | VQG | Visual | V & T |
| --- | --- | --- | --- | --- | --- | --- |
| Question | 1.71 | 3.23 | 4.41 | **8.61** | 0.76 | 1.16 |
| Response | 1.34 | – | **1.57** | – | – | 0.66 |

Table 6: Results of evaluating using multi-reference BLEU.

Table 4 presents example generations by our best performing systems. In human evaluation shown in Tables 5, the model that encodes both visual and textual context outperforms others. We note that human judges preferred the top generation in the n-best list over the reranked best, likely due to the tradeoff between a safe and generic utterance and a riskier but contentful one. The human gold references are consistently favored throughout the table. We take this as evidence that IGC$_{\text{Crowd}}$ test set provides a robust and challenging test set for benchmarking the progress on the task.

As shown in Table 6, BLEU scores are low, as is characteristic for language tasks with intrinsically diverse outputs (Li et al., 2016b,a). On BLEU, the multimodal *V&T* model outperforms all the other models across test sets, except for the VQG model which does significantly better. We attribute this to two issues: (1) the VQG training dataset partly contains event-centric images similar to our IGC$_{\text{Crowd}}$ test set, (2) Training on a high-quality crowdsourced dataset with controlled parameters can be, to a degree, more effective than training data on the wild data such as Twitter. However, crowdsourcing multi-turn conversations between paired workers at large scale is very expensive, which encourages using readily available big data in the wild.

Overall, in both automatic and human evaluation, our question generation models are more successful than response generation. This disparity may overcome by (1) developing more sophisticated systems for richer modeling of long contexts across multiple conversational turns, (2) larger high-quality training datasets.

## 8 Conclusions

We have introduced a new task of multimodal image-grounded conversation, in which, when given an image and a natural language text, the system must generate meaningful conversational turns. We are releasing to the research community a crowdsourced dataset of 4,222 high-quality multi-turn conversations about eventful images and multiple references. This dataset contains natural human-human conversations and is not tied to the characteristics of any given social media resources, e.g., Twitter or Reddit. We thus expect this shared corpus to remain stable over time. Although here we used Twitter data for training, using a variety of other training resources in future work should be illuminating.

Our experiments provide evidence that capturing multimodal context improves the quality of question and response generations. The gap between the performances of our best models and humans opens opportunities further research in the continuum from casual chit-chat conversation to more topic-oriented dialog. We expect that addition of other kinds of grounding, such as temporal and geolocation information, can further improve the performance.

# References

Stanislaw Antol, Aishwarya Agrawal, Jiasen Lu, Margaret Mitchell, Dhruv Batra, C. Lawrence Zitnick, and Devi Parikh. 2015. VQA: Visual question answering. In *International Conference on Computer Vision (ICCV)*.

Collin F. Baker, Charles J. Fillmore, and John B. Lowe. 1998. The berkeley framenet project. In *Proceedings of the 17th International Conference on Computational Linguistics - Volume 1*. Association for Computational Linguistics, Stroudsburg, PA, USA, COLING '98, pages 86–90.

Jianfu Chen, Polina Kuznetsova, David Warren, and Yejin Choi. 2015. Déjà image-captions: A corpus of expressive descriptions in repetition. In *Proceedings of the 2015 Conference of the North American Chapter of the Association for Computational Linguistics: Human Language Technologies*. Association for Computational Linguistics, Denver, Colorado, pages 504–514.

Kyunghyun Cho, Bart Van Merriënboer, Caglar Gulcehre, Dzmitry Bahdanau, Fethi Bougares, Holger Schwenk, and Yoshua Bengio. 2014. Learning phrase representations using rnn encoder-decoder for statistical machine translation. *arXiv preprint arXiv:1406.1078*.

Abhishek Das, Satwik Kottur, Khushi Gupta, Avi Singh, Deshraj Yadav, José M. F. Moura, Devi Parikh, and Dhruv Batra. 2017. Visual dialog. In *CVPR*.

Jacob Devlin, Hao Cheng, Hao Fang, Saurabh Gupta, Li Deng, Xiaodong He, Geoffrey Zweig, and Margaret Mitchell. 2015. Language models for image captioning: The quirks and what works. In *Proceedings of the 53rd Annual Meeting of the Association for Computational Linguistics and the 7th International Joint Conference on Natural Language Processing (Volume 2: Short Papers)*. Association for Computational Linguistics, Beijing, China, pages 100–105.

Jeff Donahue, Lisa Anne Hendricks, Sergio Guadarrama, Marcus Rohrbach, Subhashini Venugopalan, Kate Saenko, and Trevor Darrell. 2014. Long-term recurrent convolutional networks for visual recognition and description. *CoRR* abs/1411.4389.

Hao Fang, Saurabh Gupta, Forrest N. Iandola, Rupesh Srivastava, Li Deng, Piotr Dollár, Jianfeng Gao, Xiaodong He, Margaret Mitchell, John C. Platt, C. Lawrence Zitnick, and Geoffrey Zweig. 2014. From captions to visual concepts and back. *CoRR* abs/1411.4952.

Ali Farhadi, Mohsen Hejrati, Mohammad Amin Sadeghi, Peter Young, Cyrus Rashtchian, Julia Hockenmaier, and David Forsyth. 2010. Every picture tells a story: Generating sentences from images. In *Proceedings of the 11th European Conference on Computer Vision: Part IV*. Springer-Verlag, Berlin, Heidelberg, ECCV'10, pages 15–29.

Michel Galley, Chris Brockett, Alessandro Sordoni, Yangfeng Ji, Michael Auli, Chris Quirk, Margaret Mitchell, Jianfeng Gao, and Bill Dolan. 2015. deltaBLEU: A discriminative metric for generation tasks with intrinsically diverse targets. In *Proc. of ACL-IJCNLP*. Beijing, China, pages 445–450.

Micah Hodosh, Peter Young, and Julia Hockenmaier. 2013. Framing image description as a ranking task: Data, models and evaluation metrics. *J. Artif. Int. Res.* 47(1):853–899.

Mark Hopkins and Jonathan May. 2011. Tuning as ranking. In *Proceedings of the 2011 Conference on Empirical Methods in Natural Language Processing*. Association for Computational Linguistics, Edinburgh, Scotland, UK., pages 1352–1362.

Ting-Hao (Kenneth) Huang, Francis Ferraro, Nasrin Mostafazadeh, Ishan Misra, Aishwarya Agrawal, Jacob Devlin, Ross Girshick, Xiaodong He, Pushmeet Kohli, Dhruv Batra, C. Lawrence Zitnick, Devi Parikh, Lucy Vanderwende, Michel Galley, and Margaret Mitchell. 2016. Visual storytelling. In *Proceedings of the 2016 Conference of the North American Chapter of the Association for Computational Linguistics: Human Language Technologies*. Association for Computational Linguistics, San Diego, California, pages 1233–1239.

Jiwei Li, Michel Galley, Chris Brockett, Jianfeng Gao, and Bill Dolan. 2016a. A diversity-promoting objective function for neural conversation models. In *Proceedings of the 2016 Conference of the North American Chapter of the Association for Computational Linguistics: Human Language Technologies*. Association for Computational Linguistics, San Diego, California, pages 110–119.

Jiwei Li, Michel Galley, Chris Brockett, Jianfeng Gao, and Bill Dolan. 2016b. A persona-based neural conversation model. In *Proceedings of the 2015 Conference of the North American Chapter of the Association for Computational Linguistics: Human Language Technologies*. Association for Computational Linguistics.

Chin-Yew Lin and Franz Josef Och. 2004. Automatic evaluation of machine translation quality using longest common subsequence and skip-bigram statistics. In *Proceedings of the 42Nd Annual Meeting on Association for Computational Linguistics*. Association for Computational Linguistics, Stroudsburg, PA, USA, ACL '04.

Tsung-Yi Lin, Michael Maire, Serge Belongie, James Hays, Pietro Perona, Deva Ramanan, Piotr Dollr, and C. Lawrence Zitnick. 2014. Microsoft coco: Common objects in context. In *ECCV*. Zrich. Oral.

Chia-Wei Liu, Ryan Lowe, Iulian Serban, Mike Noseworthy, Laurent Charlin, and Joelle Pineau. 2016.

## A  IGC$_{\text{Twitter}}$ Training Dataset

Previous work in neural conversation modeling (Ritter et al., 2011; Sordoni et al., 2015) has successfully used Twitter as the source of millions of natural conversations. As training data, we sampled 250K quadruples of {visual context, textual context, question, response} tweet threads from a larger dataset of 1.4 million, extracted from the Twitter Firehose over a 3-year period beginning in May 2015 and filtered to select just those conversations in which the initial turn was associated with an image and the second turn was a question. Regular expressions were used to detect questions. To improve the likelihood that the authors are experienced Twitter conversationalists, we further limited extraction to those exchanges where users had actively engaged in at least 30 conversational exchanges during a 3-month period.

Twitter data is noisy; we performed simple normalizations, and filtered out tweets that contained mid-tweet hashtags, were longer than 80 characters[3] and contained URLs not linking to the image. Table 7 presents example conversations from this dataset. Although the filters result in significantly higher quality of the extracted conversations, issues remain. Table 8 shows two such problematic examples. In the left example, the conversation is not grounded in the image and the textual context, but rather in the participants' established relation or prior history; A random sample of tweets suggests that about 46% of the Twitter conversations is affected by prior history between users, making response generation particularly difficult. In addition, the abundance of screenshots and non-photograph graphics (such as the right example in Table 8) is potentially a major source of noise in extracting features for neural generation. Despite this, relatively large training dataset is crucial for training open-ended conversational models and the $IGC_{Twitter}$ constitutes a very large training set for the task of IGC. Furthermore, this data is relatively cheap to extract and could be used in much greater quantities. We use the IGC$_{\text{Twitter}}$ dataset as our primary training data.

---
[3] Pilot studies showed that 80 character limit more effectively retains one-sentence utterances that are to the point.

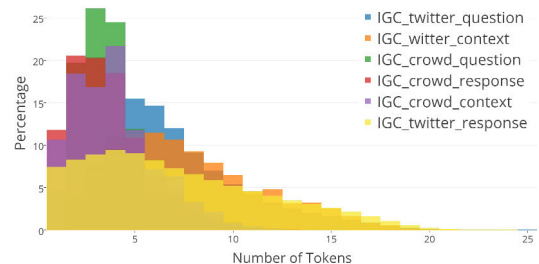

Figure 8: Distribution of the number of tokens across datasets.

## B  Data Analysis

In this Section we provide a variety of analysis, comparing the IGC dataset to the other existing vision & language datasets.

Figure 8 shows the distribution of the number of tokens per sentence. On average, the IGC$_{\text{Twitter}}$ dataset has longer sentences. Figure 9 visualizes the n-gram distribution (with n=6) of questions across datasets. IGC$_{\text{Twitter}}$ is the most diverse set, with the lighter-colored part of the circle indicating sequences with less than 0.1% representation in the dataset.

Figure 10 compares IGC questions with VQG (Mostafazadeh et al., 2016b) and VQA (Antol et al., 2015) questions in terms of vocabulary size, percentage of abstract terms, and inter-annotation textual similarity. The COCO (Lin et al., 2014) image captioning dataset is also included as a point of reference. The IGC$_{\text{Twitter}}$ dataset has by far the largest vocabulary size, making it a more challenging dataset for training purposes. The IGC$_{\text{Crowd}}$ and IGC$_{\text{Twitter}}$, in order, have the highest ratio of abstract to concrete terms. Broadly, abstract terms refer to intangibles, such as concepts, qualities, and feelings, whereas concrete terms refer to things that can be experienced with the five senses. It appears that conversational content may often involve more abstract concepts than either captions or questions targeting visible image content.

| | | | | |
|---|---|---|---|---|
| **Visual Context** | 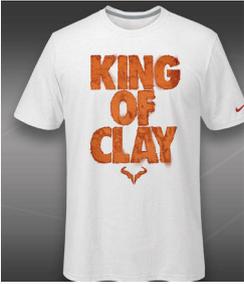 | 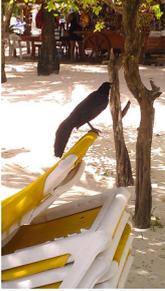 | 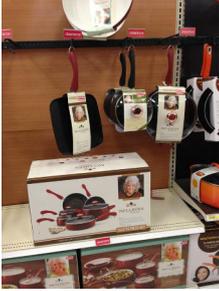 | 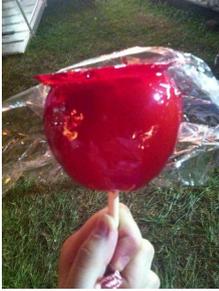 |
| **Textual Context** | Oh my gosh, i'm so buying this shirt. | I found a cawaii bird. | Stocking up!! | Only reason I come to carnival. |
| **Question** | Where did you see this for sale? | Are you going to collect some feathers? | Ayee! what the prices looking like? | Oh my God. How the hell do you even eat that? |
| **Response** | Midwest sports | There are so many crows here I'd be surprised if I never found one. | Only like 10-20% off..I think I'm gonna wait a little longer. | They are the greatest things ever chan. I could eat 5! |

Table 7: Example conversations in the IGC$_{\text{Twitter}}$ dataset.

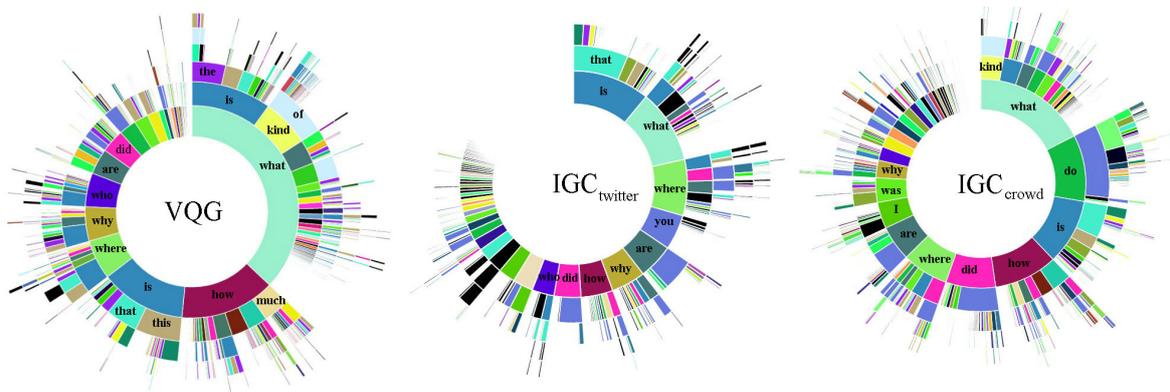

Figure 9: Distributions of n-gram sequences in questions in VQG, IGC$_{\text{Twitter}}$, and IGC$_{\text{Crowd}}$.

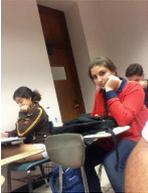

| | |
|---|---|
| Smile. | What's your excuse? |
| Why are you so obsessed with me? | Nca nationals? which day? |
| Oh pls | Day 2 i believe ! if you go on youtube it should be the first one ! |

Table 8: Example Twitter conversations that add noise to the dataset.

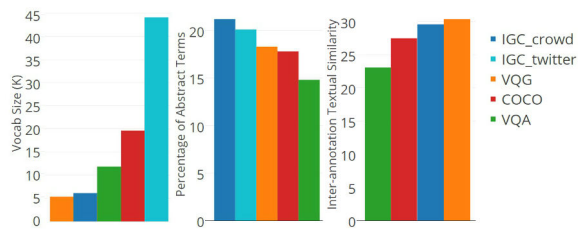

Figure 10: Comparison of V&L datasets.

It has been shown that humans achieve greater consensus on what a natural question to ask given an image (the task of VQG) than on captioning or asking a visually verifiable question (VQA) (Mostafazadeh et al., 2016b). The right-most plot in Figure 10 compares the inter-annotation textual similarity of our IGC$_{\text{Crowd}}$ questions using a smoothed BLEU metric (Lin and Och, 2004). IGC$_{\text{Twitter}}$ is excluded from this analysis as the data is not multireference. Contextually grounded questions of IGC$_{\text{Crowd}}$ are competitive with VQG in inter-annotation similarity.